\documentclass[twoside,leqno,twocolumn]{article}

\usepackage[letterpaper]{geometry}

\usepackage{ltexpprt}
\usepackage{graphicx}
\usepackage{multicol}
\usepackage{expl3}
\usepackage{url}
\newcommand{\universityname}{University of Bath, Bath, UK: \hfil\break{\tt \{ag2214,nkdf20,np700,masjhd,asn42\}@bath.ac.uk}}

\begin{document}

\title{\Large The Influence of Domain-Based Preprocessing on Subject-Specific Clustering}
\author{Alexandra Gkolia\thanks{\universityname.}
\and Nikhil Fernandes\footnotemark[1]
\and Nicolas Pizzo\footnotemark[1]
\and James Davenport\footnotemark[1]
\and Akshar Nair\footnotemark[1]
}

\date{}

\maketitle


\fancyfoot[R]{\scriptsize{Copyright \textcopyright\ 2020 by SIAM\\
Unauthorized reproduction of this article is prohibited}}





\begin{abstract} \small\baselineskip=9pt
	The sudden change of moving the majority of teaching online at Universities due to the global Covid-19 pandemic has caused an increased amount of workload for academics. One of the contributing factors is answering a high volume of queries coming from students. As these queries are not limited to the synchronous time frame of a lecture, there is a high chance of many of them being related or even equivalent. One way to deal with this problem is to cluster these questions depending on their topic. In our previous work, we aimed to find an improved method of clustering that would give us a high efficiency, using a recurring LDA model. Our data set contained questions posted online from a Computer Science course at the University of Bath. A significant number of these questions contained code excerpts, which we found caused a problem in clustering, as certain terms were being considered as common words in the English language and not being recognised as specific code terms. To address this, we implemented tagging of these technical terms using Python, as part of preprocessing the data set. In this paper, we explore the realms of tagging data sets, focusing on identifying code excerpts and providing empirical results in order to justify our reasoning.
\end{abstract}

\section{Introduction}
A popular approach for clustering question-answer pairs uses unsupervised machine learning algorithms, such as the Latent Dirichlet Allocation (LDA) model in \cite{Goppetal19}. Due to a sudden demand for online courses at Universities, there has been an increase in workload for academics. For example, the Computer Science department at the University of Bath has introduced five new MSc courses for the academic year 2020-21, which has resulted in an influx of over 200 students. As these courses are structured to be completely online, the lectures are pre-recorded and available in video on demand format. Given that there are no set hours to watch lectures, student queries are dealt with in a question-answer format within community-based websites and applications such as Microsoft Teams, Google Groups, Blackboards etc.

\subsection{Domain-specific words}
Our previous research \cite{NANJA29} focused on finding an efficient way to cluster question-answer banks by recursively applying the LDA model. The data set was obtained from the Data Structures and Algorithms course offered at the University of Bath. On inspection of the data set, we noticed that there were terms recurring throughout different questions which were domain-specific. One the main areas these words were related to was code excerpts. During our analysis of the recursive model, we further noticed that that the existing preprocessing infrastructure prior to implementing LDA considered these terms as common English words (and indeed often stop words) and not possible candidates for topic keywords.

To counter this, we manually chose the words that were tagged and set a procedure to automatically tag them within the data set, then proceeded with the recursive model \cite{NANJA29}. This gave us a significantly better clustering of the data set. In this paper, we investigate the effects of tagging our data set. We mainly focus on tagging specific terms such as code words, and mathematical terms such as 'O' (in actuality $\mathcal{O}$, used to denote asymptotic complexity) and 'modulo'.
 
\subsection{LDA input parameters}
Another problem faced during implementation of the recursive model was optimising the hyperparameters. As described later in Section \ref{sec:HDPestimator}, one can exploit the non-parametric nature of HDP and use it to obtain an estimate for the optimal number of topics the data should be clustered into. This estimate is further used as an input for the LDA model. However, we choose to use two recursions of the HDP model. At first glance, recursion seemed to provide a good estimate, however further recursion suffered from Zeno's paradox and gave incorrect values. We provide further results within the paper to describe why this is the case.

\section{Latent Dirichlet Allocation}\label{sec:LDA}
One of the most commonly used data clustering techniques is Latent Dirichlet Allocation (LDA) \cite{lda2003}, a statistical method where each data point is assigned a probability of it belonging to a certain topic. Statistical techniques are useful in machine learning and data clustering as they allow for multiple clustering possibilities, as data points aren’t clustered to topics definitively. LDA relies on the Dirichlet distribution to extract these topic probabilities. The LDA model applied to document clustering is laid out as follows, with some key definitions. Document refers to a collection of words and corpus refers to a collection of documents. Topic refers to an unobserved ‘latent’ group generated by the LDA model.

\begin{figure}
	\includegraphics[width=\columnwidth]{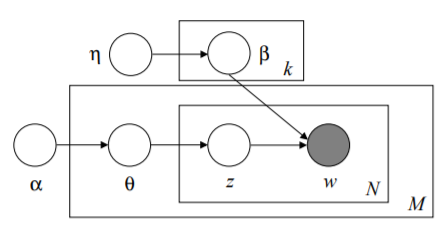}
	\caption{Plate Notation for the LDA model  \cite{lda2003}. Here, $M$ denotes a document, $N$ denotes the words contained within the document, $k$ denotes the identified topics, and $z$ denotes the topic(s) assigned to the observed word $w$. Each rectangle denotes a repetition within the model to form the overall structure of the corpus data. Variables $\beta$ and $\theta$ denote the word-topic and topic-document distributions.}
	\label{fig:LDAPlateNotation}	
\end{figure}

Viewed in Figure \ref{fig:LDAPlateNotation}, the LDA model takes two input parameters to determine the distribution variables:
\begin{itemize}
	\item $\beta_k$ is the word distribution for a given topic $k$ and is determined by $\beta_k \sim Dir(\eta)$, where $\eta \in [ 0, \infty)$ is the input parameter
	\item $\theta_i$ is the topic distribution for a given document $i$, determines by $\theta_i \sim Dir(\alpha)$, where $\alpha \in [ 0, \infty)$ is the input parameter.
\end{itemize}
These distributions, $\beta_k$ and $\theta_i$, can be viewed as two matrices, on which the relationships between words, documents and topics are created. LDA forms the assumption that these variables are generated by underlying probability distributions, specifically the symmetric Dirichlet distribution. As these distributions rely on the input parameters $\alpha$ and $\eta$, these can be adjusted to determine the 'mixture' of topics in a document as well as the 'mixture' of words in a topic.

LDA presents a key advantage for use in text clustering. As it is a generative statistical model, it automatically produces latent groups for the data, which can be considered without conflict as topics or clusters. However, LDA presents certain limitations to a data clustering model. By requiring a predetermined parameter for the number of topics, LDA is susceptible to producing inefficient clusterings, where the data is clustered into too few or too many topics. Thus it is important to determine an optimum topic number prior to using the LDA model. This can be hard to interpret with a large, complex data set.
An important tool used in our methodology was the Hierarchical Dirichlet Process (HDP), a generalisation of LDA. We used this model as a optimal topic number estimator, as it does not require the setting of number of topics parameter as in LDA. HDP is based on the Dirichlet Process, which is the infinite-dimensional generalization of the Dirichlet distribution.

\subsection{Examples of Implementation} Pritchard et al. \cite{Pritchard945} first introduced the notion of using the LDA model to observe the genetic distribution of a population. They aimed to gain an understanding of the population structure by assigning individuals to groups based on their genotypes. Here are a few examples of different applications of the LDA model:

\begin{itemize}
	\item Cybersecurity Strategy Analysis (Kolini et al.\cite{kolinetal17}): The authors used the LDA model to cluster the National Cybersecurity Strategies (NCS) documents from 60 countries. They tried to find similarities between the countries depending on certain keywords assigned to their respective NCSs. One interesting observation they made was the distinct similarity between documents developed by countries in the EU and the countries part of NATO.
	\item Consumer Financial Protection Bureau (Bastani et al.\cite{Batsanietal18}): Their model used the LDA algorithm to extract latent topics from Consumer Financial Protection Bureau (CFPB) consumer complaints. This helped them monitor the effectiveness of new regulations being set by the CFPB. They specified that this method was not a replacement for human analysis but an assistive tool to increase efficiency.	
	\item Extracting information from Twitter (Gemci et al.\cite{Gemcietal13}, Montenegro et al.\cite{Montenegroetal18}): There have been various attempts to extract information from social media platforms such as Twitter. The authors of \cite{Montenegroetal18} used the LDA model combined with support vector machines to to perform sentiment analysis. Another application regarding Twitter, was done by \cite{Gemcietal13} to analyse Turkish tweets. This included the challenging task of modifying the preprocessing used for LDA to account for the change in language.
	\item Book Recommendation System (Alharthi et al.\cite{Alharthi2017}): A data mining approach to use the LDA model on a users social network data to extract key terms which would suggest book recommendations. 
	\item Fraud Detection (Wang et al.\cite{WANG18}): Uses the LDA model to extract text features from accident insurance claims. These text features were fed to a deep neural networks as training data, for detecting fraudulent insurance claims.  
\end{itemize}

\section{Hierarchical Dirichlet Process}\label{sec:HDP}
The Hierarchical Dirichlet Process (HDP) is a Bayesian clustering model for grouped data. It is similar to the LDA model, however requires no preset parameters, thus it can be viewed as a generalisation of the LDA model for the purposes of our implementation, see \cite{HDP2006}.

Indeed, HDP is constructed in the same way as LDA, considering documents to be a mixture of latent topics, where the latent topics are mixtures of words. The model establishes the relationship between words, topics and documents using the following three variables, represented as matrices:
\begin{itemize}
	\item $\alpha_k$ representing top level Dirichlet variables sampled for a given topic $k$
	\item $\theta_j$ representing the topic distribution for a given document $j$
	\item $\varphi_k$ representing the word distribution for a given topic $k$
\end{itemize}

To generate the matrices for these variables, the HDP algorithm assumes the variables are modelled by the symmetric Dirichlet distribution, with different parameters in each case to reflect variable dependencies. The matrices are constructed as follows, with parameters $\beta, \gamma, \eta$:

$$\alpha_k \sim Dir(\gamma /K)$$
$$\theta_j \sim Dir(\eta \, \alpha_k)$$
$$\varphi_k \sim Dir(\beta)$$
$$z_{ij} \sim \theta_j, \, \, x_{ij} \sim \varphi_{z_{ik}}$$

\vspace{0.5em}

In the HDP model above, K denotes the number of topics and is assumed to be infinite, as all possible topics are considered. The variables $z_{ij}$ and $x_{ij}$ are sampled from $\theta_j$ and $\varphi_{z_{ik}}$ respectively.

\begin{figure}
	\begin{center}
	\includegraphics[width=5cm]{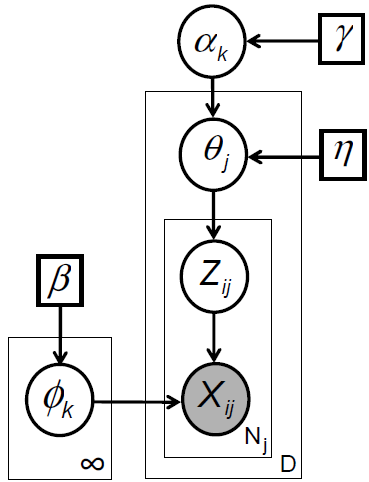}
	\end{center}
	\caption{Plate Notation for the HDP model \cite{Newman2009}. Here, $D$ signifies the number of documents, $N$ signifies the number of words contained within the document, $K$ denotes the number of topics, and $z_{ij}$ denotes the topic(s) assigned to the observed word $x_{ij}$, and $\phi_k$ represents the probability of a word given a topic $k$. The rectangles present in the diagram denote a repetition within the model to form the overall structure of the corpus data.}
	\label{fig:HDPPlateNotation}	
\end{figure}

The inherent difference between the LDA and HDP models is that the former requires setting of a predetermined number of topics $K$ into which to cluster the data. This is in contrast to HDP, where the value for the number of topics is viewed as a variable generated by a Dirichlet Process. The additional layer of the Dirichlet Process is added to the HDP model, thus the model is formed in a 'hierarchical' manner.

\subsection{Predicting an optimum topic number}\label{sec:HDPestimator}

The non-parametric nature of HDP is one of its key advantages over LDA, which is the reason we use it to determine an optimum topic number for input into our LDA clustering model. LDA is difficult to implement without full knowledge of the data set, as the number of topics has to be determined. This problem is solved by using HDP, which takes into consideration an infinite number of topics and assigns a probability of relevance to each of these. When setting a criterion, the most relevant topics remain, so this number can be taken as an estimate of the number of topics. In the first run of HDP, the criterion is taken as $\frac{1}{n}$, where $n$ is the size of the corpus set. As experimentally this prediction proves to be an overestimate, a second run of HDP is performed with the criterion set to $\frac{1}{x}$, where $x$ is the number of topics given by the first HDP estimate. These two runs of HDP are referred to throughout as HDP-1 for the first run and HDP-2 for the second run.

\section{Domain-specific Tagging}\label{sec:DataTagging}
General preprocessing tools are already available to us for the purposes of preparing data for Natural Language Processing. However, these are often tools such as punctuation and stop word removal, which concentrate on reducing data into simpler forms. Rarely do these methods effectively extract important meanings from text; methods such as stemming attempt to deduce the root form of a word, but do not take context into account. To separate meaning from text, especially for highly subject-specific or technical terms, these methods often fall short of extracting significant words. We expand more on this topic in Section \ref{sec:Methodology}.

There is no inbuilt method for detecting coding language terminology and tagging these keywords within data is not a widely available natural language processing tool. In our case, dealing with a data set containing code excerpts, this is significant, as coding terms such as 'if', 'for' and 'while' will be removed, resulting in a change in the meaning of our data. This tagging method is specific to our data set, but can be adapted to suit other topics where certain keywords are not recognised as significant words in preprocessing. Automatic tagging of data is not currently available within the data preprocessing tools offered for natural language processing.

Our method of domain-specific tagging is performed semi-automatically; tags of a predefined list of significant words are created, avoiding the issue of these words being removed by preprocessing tools. Additionally, the tags are treated as regular feature words by the clustering model and contribute to the final clustering. This is in contrast to the untagged data, where these significant words were removed and did not contribute to the final clustering.

The keywords we tagged were specific to our data set, which consisted of computer science related questions. This use of tagging and the list of tags can be adapted to other subjects by manually selecting significant words which are otherwise removed as stop words, or are not recognized by the relevant dictionary in preprocessing. Thus, this method can be extended to various data sets, including subject-specific and technical data sets.

\section{Our Methodology}\label{sec:Methodology}
In order to investigate the effect of data tagging on data clustering, we implemented the LDA model as detailed in Section \ref{sec:LDA}. We used the Python package 'gensim' which offered us an implementation of LDA with various input parameters. The main input parameter of LDA is the number of topics (or clusters) the model should cluster to. In order to find the optimal number of topics to cluster to, we used a HDP topic estimator \cite{NANJA29}.

\subsection{Data preprocessing}

Our data set was comprised of around 1300 questions submitted by students attending the CM20254 Data Structures and Algorithms course at the University of Bath. This data set included the questions along with their answers and explanations, and was obtained in .txt file format. Various data preprocessing tools played an essential part in preparing our data for clustering using our model, as the original format of the data could not be easily manipulated within Python.

The first step in preprocessing was importing the text file into a relevant data structure in order to retain its properties and allow for the extraction of questions for input into the clustering model. This was achieved using a script which converted .txt files to .csv files. As our model implementations were executed in Python, we used the 'pandas' package in order to import our .csv data file into the versatile pandas 'DataFrame' structure, which allowed for greater manipulation of the data within the Python notebook.

\subsection{Data tagging method}

Tagging was achieved by inspecting the data and adding tags to the beginning of the questions, as a fully automated method was not available to us. For our data set, it was crucial to tag words which were being removed by the stop word removal tools, yet held a meaning significant in the context of a course within the field of Computer Science. The key words which were relevant to our purpose and were tagged were:
\begin{multicols}{2}
\begin{itemize}
	\item bigO
	\item modulo
	\item for
	\item if
	\item while
	\item else
	\item elseif
	\item print
\end{itemize}
\end{multicols}
As the data tagging process was performed, tags were only added for the words in context, rather than for every instance were the word would appear. For example, the words found within code, such as 'for' or 'if', were only tagged when they were found within a code excerpt. Furthermore, different forms of a concept were tagged under one word for clustering simplicity. For example, the 'Modulo' tag was applied to all questions which mentioned 'mod', 'modulo' or '\%', within context. The next step was to implement punctuation and stop word removal, which was done simply using the Natural Language Toolkit ('nltk') Python package.

After the preprocessing stage, we produced five random permutations of our data set in order to view any differences in clustering caused by a different ordering of the data. Subsequently, the original and tagged data sets, as well as their five permutations, became the input for our model. Our model made use of an HDP estimator to predetermine the optimal number of topics needed to cluster the data set. Two estimates were used from the HDP estimator and the LDA model was implemented with both to cluster our datasets.

\section{Empirical Results}\label{sec:Results}
In our methodology, we performed runs of LDA using an HDP topic estimator on both the original data set and the tagged data set, as well as the five different permutations of each. We evaluated the output files in order to establish differences between the tagged and untagged data sets, as well as compare the efficiency between HDP-1 and HDP-2, to determine which performs better as a topic number estimator. Thus, this section is split into two main subsections: the first aims to explore the differences between HDP-1 and HDP-2 as topic estimators, while the second investigates tagging as a method of improving data classifications when clustering. For clarity, individual clusters are referred to as topics throughout this section, and topics of untagged and tagged data outputs are referred to as U-$x$ and T-$x$ respectively, where $x$ is the topic number as determined by the model. We also refer to outputs as untagged or tagged clusterings if obtained from the untagged or tagged data sets respectively.

\subsection{Results of Iterating HDP}
\label{sec:IteratingHDPResults}

\begin{figure}
	\includegraphics[width=\columnwidth]{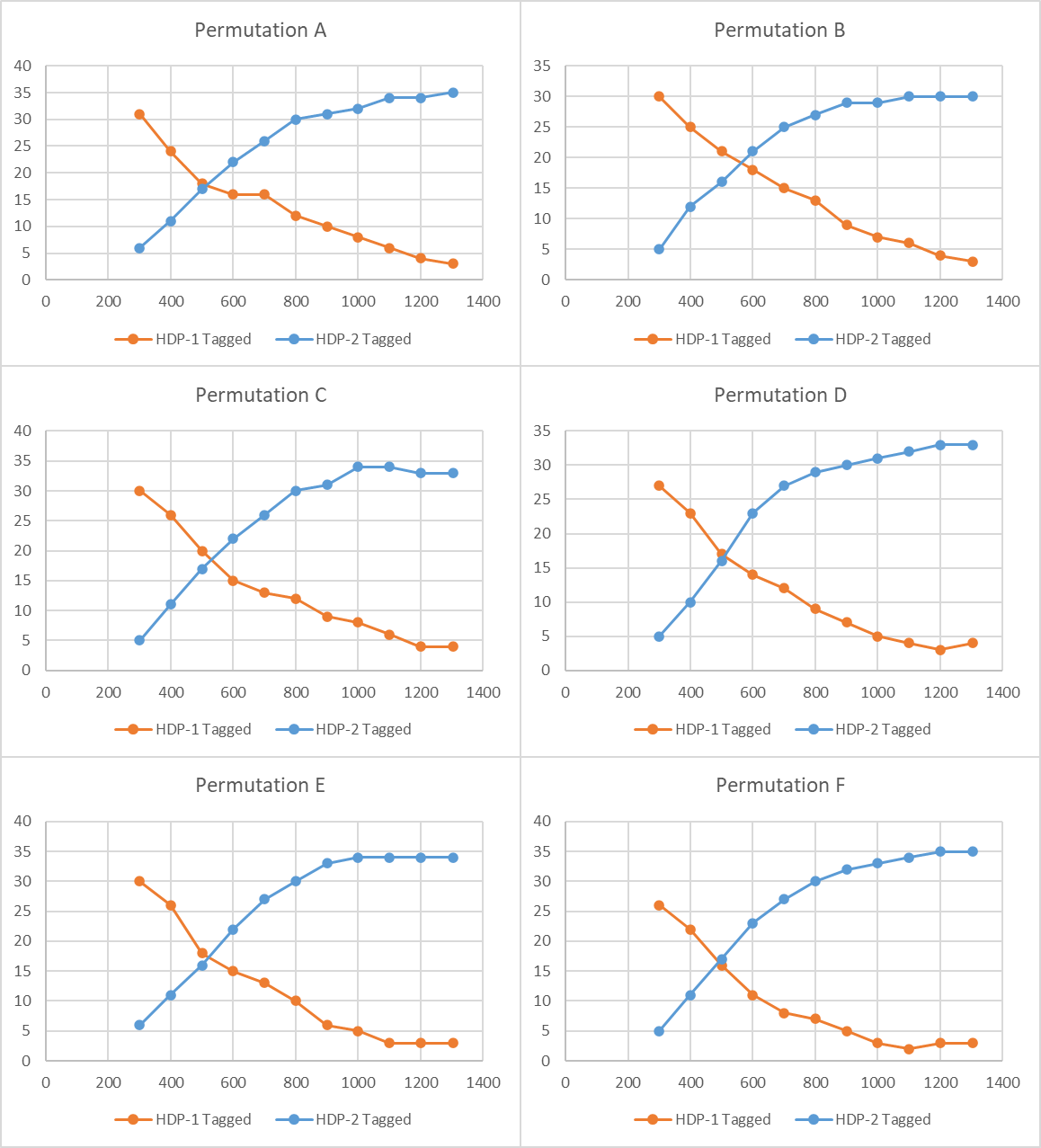}
	\caption{Number of topics used from HDP-1 and HDP-2 estimates against the number of questions for the tagged data set}
	\label{fig:TaggedTopicsHDP1vs2}
\end{figure}

In our previous research \cite{NANJA29}, we found that HDP-2 performs better than HDP-1 when clustering large data sets. In an attempt to verify this for our tagged data set, we performed runs of both the HDP-1 and HDP-2 estimates, retaining the data for topics which were being clustered to. This data is displayed in Figure \ref{fig:TaggedTopicsHDP1vs2}, where we observe that for a small number of questions, HDP-1 is making use of a large number of topics, but for large questions, HDP-1 clusters the data into significantly fewer topics. In contrast, the HDP-2 estimate causes the data to be clustered into a larger number of topics as the number of questions within the data set increases.

\begin{figure}
	\includegraphics[width=\columnwidth]{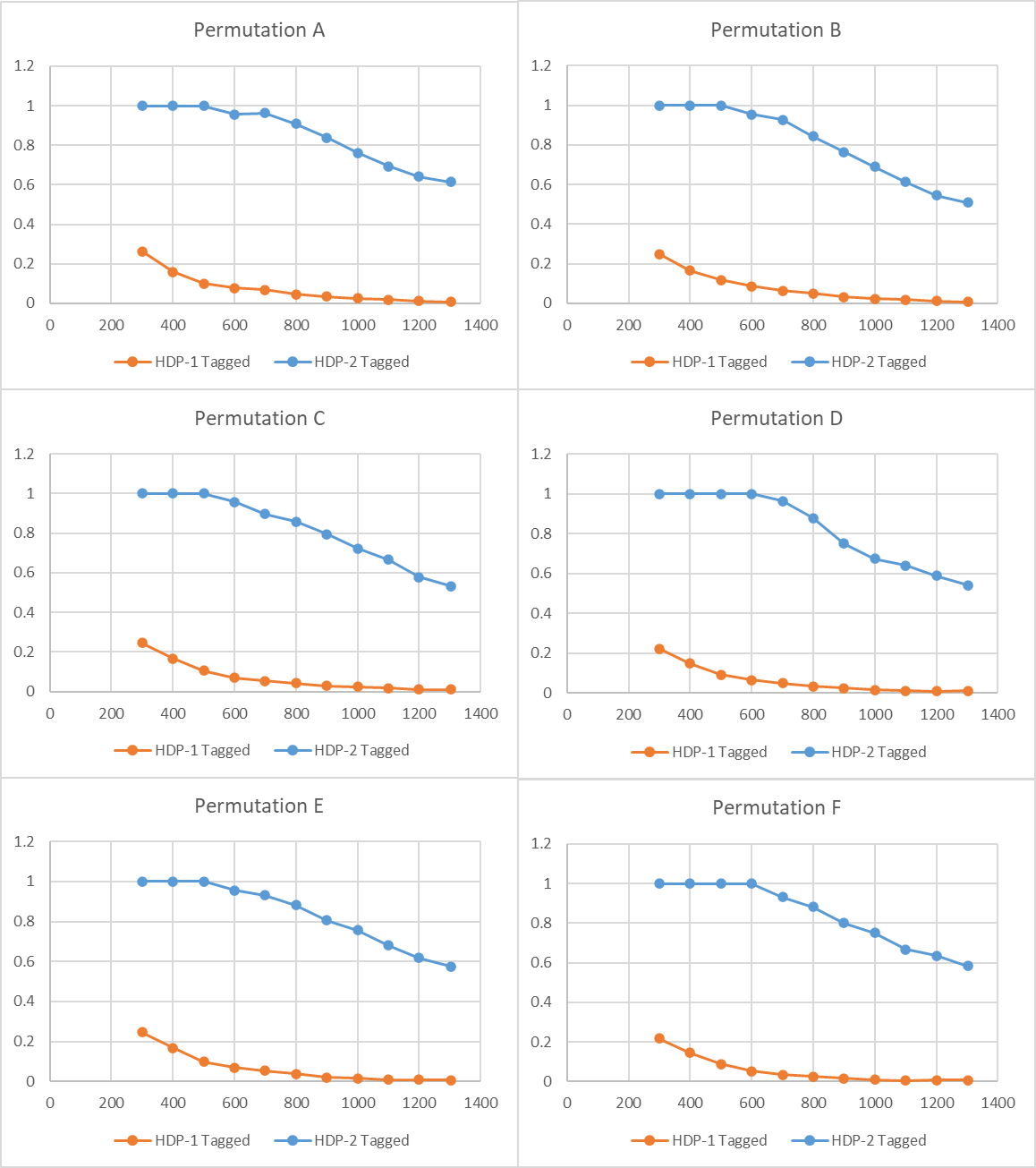}
	\caption{Efficiency ratio of HDP-1 and HDP-2 estimates against the number of questions for the tagged data set}
	\label{fig:TaggedRatioHDP1vs2}	
\end{figure}

The difference between the two estimates is further highlighted when looking at the efficiency ratio of these estimates, as shown in Figure \ref{fig:TaggedRatioHDP1vs2}. The efficiency ratio for HDP-2 is particularly high, consistently over 0.5 for the whole data set. On the other hand, the HDP-1 estimate is seen to perform poorly, and the efficiency ratio is shown to tend towards zero for large data sets. These graphs highlight that, regardless of the ordering of the data, HDP-2 always produces a more optimised estimate than HDP-1, and this is especially significant for larger data sets.
\linebreak

\subsection{Difference of tagging data}

We analysed the outputs generated by the HDP-2 estimate, which yielded interesting results. Initially, when the untagged data set was clustered using the LDA model, the HDP-2 estimator suggested 65 topics, of which 10 were of a high enough frequency to be deemed significant. On the other hand, for the tagged dataset the second run of the HDP topic estimator suggested 55 topics. Out of these, 14 were particularly significant with high frequencies of questions. The high ratio of significant topics to the estimated topics for the tagged data set would suggest an improvement in clustering.

In order to evaluate the effects of tagging on our clustering model, we qualitatively assessed the output from our model to identify differences in the clustering of the tagged and untagged data sets. Note that topic numbers were not preserved when applying our model. Our main aim was to analyse how topics were being merged and separated, focusing on the keywords associated with each topic. Here we have listed a key number of our observations that show clear justification to the use of our model.

\subsubsection{Hashtable cluster}

In both the tagged and untagged clusterings, two topics were identified in each clustering containing questions on hashtables. These two topics in the untagged data set were U-11 and U-62. Topic U-11 mainly focused on questions of hashtable operations and functions, while U-62 predominantly contained questions on open and closed hastables.

Within the tagged output, the two topics were T-10 and T-48. Topic T-10 contained questions concerning open hashtables, including questions on buckets and steps required for performing searches. Topic T-48 focuses on closed hashtables, including questions concentrating on operations such as insert, as well as calculations of the hash function (which uses modulo arithmetic, thus modulo is identified as a keyword). This is significant as modulo was one of the tagged words implemented within our model. Our clustering separated questions on definitions of hashtables and those on the operations of hashtables into two clusters. This shows a significant improvement in clustering, by firstly isolating the types of questions and then the topics of questions.

\subsubsection{Cluster containing code excerpts}

Another interesting set of keywords belonged to the domain of code terminology. In the untagged output, topic U-64 mainly contains questions which include code excerpts. This topic is closely related to topic T-39 in the tagged data clustering, which identifies most of the same keywords, along with the tagged words ‘for’, ‘if’, ‘print’ and ‘while’, which had not been previously identified prior to tagging. There was a substantial improvement seen by tagging these code words; the untagged cluster contained some irrelevant questions, namely Q645, Q759, Q769, and Q1125, which are filtered out in the tagged version. Also, out of the 19 additional questions which were added to T-39 (and did not appear in U-64), 18 contained code excerpts, indicating that tagging makes it easier to identify and group together code excerpts by correctly identifying code words and using them for effective clustering. Further establishing our claim to an improved clustering, tagging is effective in identifying and prioritising the clustering of the type of questions, in this case questions containing code excerpts.

\begin{figure}
	\includegraphics[width=\columnwidth]{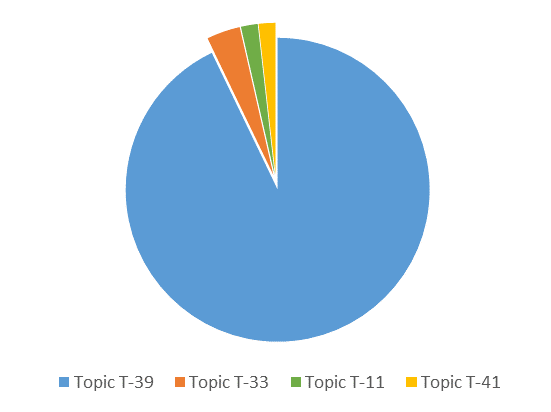}
	\caption{Distribution of Questions from Topic U-64 after Tagging}
	\label{fig:PieChartCoding}	
\end{figure}

To provide an example of how questions are distributed to different topics after tagging, Figure \ref{fig:PieChartCoding} represents the topics allocated to the questions in U-64. As can be observed, most questions were assigned to T-39, with the 4 unrelated questions being assigned to different topics, as mentioned previously.

\subsubsection{'BigO' tag and the complexity cluster}

Tagging ‘BigO’ as a keyword caused a significant difference in the clustering of questions concerning Big-O terminology and the calculation of asymptotic complexity. In the untagged data clustering, there were two main topics within this group. Topic U-4 contained questions on runtimes and the quality of different complexity approximations. This topic also contained questions on insertion sort and its complexity analysis. Similarly, Topic U-49 contained questions concerning asymptotic time complexities of algorithms, and further included questions on the complexity of functions.
\linebreak

In the tagged data clustering, there were three relevant topics identified which contained questions regarding BigO notation and complexity, namely topics T-20, T-25 and T-37. Topic T-20 contains a large proportion of complexity questions, also tagged by the 'BigO' tag. Most questions in the cluster are about runtime and the definitions of Big-O, -Theta and -Omega, as well as the complexity of functions. It also contains Big-O related questions which do not appear in topic U-49, due to the tag 'BigO'. This is in contrast to topic T-25, which also contains the keywords complexity and time, however focuses on more applied questions concerning best, average, bad or good runtimes. The third topic, Topic T-37, contains questions about the theoretical properties of Big-O and -Theta complexities, like transitivity, as well as questions on Dijkstra’s algorithm. The 'BigO' tag was recognised as ‘big’, and most questions tagged with 'BigO' appeared in this topic. Q1053 did not relate to the keywords in this cluster, but was included since it contained the word ‘biggest’, converted to big and confused with 'BigO'. Thus, there was an observed improvement overall in the identification of Big-O questions using the ‘BigO’ tag; more questions were deemed significant and clustered into their relevant clusters after tagging.

\subsubsection{Further significant clusters}

Other clusters in the final groupings weren't affected significantly by tagging. Tagging was mostly relevant to the certain words that were not being recognised previously, and impacted the relevant clusters concerning those words. The rest of the clusters formed from the data remained fairly stable, a testament to the reliability and effectiveness of LDA as a clustering method for most text data.

For example, the cluster on sorting algorithms remained mainly unchanged in both clusterings. In the untagged data clustering, topic U-6 contains questions on various sorting algorithms, including quicksort and bubble sort, concerning their stability as well as comparisons between them. This topic is similar to topic T-26 in the tagged data set, which identifies most of the same keywords and also concerns stabilities of sorts. These topics contained a majority of similar questions, with 84.4\% overall similarity, of which most of the differences were due to 46 more questions being added to the cluster after tagging.

Another example is the cluster on trees. In the untagged clustering, topic U-54 is the main topic containing questions regarding trees, including binary trees, AVL trees and finding nodes using traversals, as well as balancing an AVL tree. This topic is closely related to topic T-41 in the tagged data clustering, with 201 out of the 286 questions in topic U-54 appearing in topic T-41. However, in the tagged output, AVL tree questions mainly appear in a separate topic, T-46. There are still some questions on AVL trees (Q565 and Q1129) within topic T-41, due to them containing other relevant keywords, like binary and traversal. Topic T-46 is the topic concerning AVL trees and whether they are valid trees, or balanced. Thus, after tagging, AVL tree questions are being distinguished from other tree questions, as opposed to all of them being clustered together.

\section{Conclusion and Further Work}
\subsection{Domain-specific words} We focused our attention on keywords such as 'for', 'if','else' which were part of code excerpts and terms such as 'bigO', 'modulo' which were specific words in mathematics. The reason for this approach, is that we noticed the LDA model found it difficult to distinguish between semantically significant code terms and (often the same) words within the English language with no significant meaning. Our results showed that applying this tagging process significantly improved the type of clustering. We do acknowledge that the quality of clustering is subjective and that it can be viewed with a different perspective. Our goal was to cluster these queries based on type, focusing on isolating the questions related to code, maths and then topic-based questions. 

\subsection{LDA input parameters} One of the main parts of finding an efficient clustering using the LDA model is rooted within the initial steps. Given that the LDA model needs a predetermined input number of topics, it is vital to ascertain an optimal value. Previous research suggests that, for large text documents, the HDP model provides a good approximation to this. However, our data set consists of short text pairs, which resulted in the HDP model not providing a good estimate for the number of topics when used once. On further inspection, we found that the HDP applied recursively twice, or HDP-2, satisfied the required conditions to be considered a good estimate. Further recursion of the HDP model results in Zeno's paradox by giving the effective number of topics as 1.

Furthermore, we notice that on choosing a small proportion of our data set, HDP-1 does provide a good estimate, but this changes as soon as we increase the data set to contain more than 400 questions. Given that it is fair to assume that this type of question banks will contain a large number of data points, HDP-2 is the ideal choice to predict the estimated number of topics for a clustering.

In our future work, we are looking into two avenues; the first being understanding and exploring the isolation of code excerpts from questions of varied lengths. The second would be to further explore the effects of tagging related to specific domains (focusing on the legal sector) on the clustering of large data sets.

For more details on the data set and algorithm used, please contact James Davenport and Akshar Nair.

\bibliographystyle{aaai21}
\bibliography{main_bib}

\begin{thebibliography}{10}
\providecommand{\url}[1]{\texttt{#1}}
\providecommand{\urlprefix}{URL }
\providecommand{\doi}[1]{https://doi.org/#1}

\bibitem{Alharthi2017}
Alharthi, H., Inkpen, D., Szpakowicz, S.: {Unsupervised Topic Modelling in a
  Book Recommender System for New Users}. In: eCOM@SIGIR (2017),
  \url{http://ceur-ws.org/Vol-2311/paper_4.pdf}

\bibitem{Batsanietal18}
Bastani, K., Namavari, H., Shaffer, J.: {Latent Dirichlet allocation (LDA) for
  topic modeling of the CFPB consumer complaints}. Expert Systems with
  Applications  \textbf{127},  256--271 (2019).
  \doi{]10.1016/j.eswa.2019.03.001},
  \url{http://www.sciencedirect.com/science/article/pii/S095741741930154X}

\bibitem{lda2003}
Blei, D.M., Ng, A.Y., Jordan, M.I.: {Latent Dirichlet Allocation}. J. Mach.
  Learn. Res.  \textbf{3},  993–1022 (Mar 2003)

\bibitem{NANJA29}
Fernandes, N., Gkolia, A., Pizzo, N., Davenport, J., Nair, A.: {Unification of
  HDP and LDA Models for Optimal Topic Clustering of Subject Specific Question
  Banks} (2020)

\bibitem{Gemcietal13}
{Gemci}, F., {Peker}, K.A.: {Extracting Turkish tweet topics using LDA}. In:
  2013 8th International Conference on Electrical and Electronics Engineering
  (ELECO). pp. 531--534 (2013)

\bibitem{Goppetal19}
Gropp, C., Herzog, A., Safro, I., Wilson, P., Apon, A.: {Clustered Latent
  Dirichlet Allocation for Scientific Discovery}. In: 2019 IEEE International
  Conference on Big Data. pp. 4503--4511 (12 2019).
  \doi{10.1109/BigData47090.2019.9005964}

\bibitem{kolinetal17}
Kolini, F., Janczewski, L.: {Clustering and Topic Modelling: A New Approach for
  Analysis of National Cyber Security Strategies}. In: PACIS 2017 Proceedings.
  vol.~126 (2017), \url{https://aisel.aisnet.org/pacis2017/126}

\bibitem{Montenegroetal18}
Montenegro, C., Ligutom~III, C., Orio, J., Ramacho, D.: {Using Latent Dirichlet
  Allocation for Topic Modeling and Document Clustering of Dumaguete City
  Twitter Dataset}. In: ICCDE 2018: Proceedings of the 2018 International
  Conference on Computing and Data Engineering. pp.~1--5 (05 2018).
  \doi{10.1145/3219788.3219799}

\bibitem{Newman2009}
Newman, D., Asuncion, A., Smyth, P., Welling, M.: Distributed algorithms for
  topic models. Journal of Machine Learning Research  \textbf{10},  1801--1828
  (08 2009). \doi{10.1145/1577069.1755845}

\bibitem{Pritchard945}
Pritchard, J.K., Stephens, M., Donnelly, P.: Inference of population structure
  using multilocus genotype data. Genetics  \textbf{155}(2),  945--959 (2000),
  \url{https://www.genetics.org/content/155/2/945}

\bibitem{HDP2006}
Teh, Y.W., Jordan, M.I., Beal, M.J., Blei, D.M.: {Hierarchical Dirichlet
  Processes}. Journal of the American Statistical Association
  \textbf{101}(476),  1566--1581 (2006). \doi{10.1198/016214506000000302},
  \url{https://doi.org/10.1198/016214506000000302}

\bibitem{WANG18}
Wang, Y., Xu, W.: {Leveraging deep learning with LDA-based text analytics to
  detect automobile insurance fraud}. Decision Support Systems  \textbf{105},
  87 -- 95 (2018). \doi{10.1016/j.dss.2017.11.001},
  \url{http://www.sciencedirect.com/science/article/pii/S0167923617302130}

\end{thebibliography}


\end{document}